\documentclass[letterpaper, 10 pt, conference]{ieeeconf}  

\IEEEoverridecommandlockouts                              
                                                          
\usepackage{cite}
\usepackage{amsmath,amssymb,amsfonts}
\usepackage{graphicx}
\usepackage{textcomp}
\usepackage{makecell}
\usepackage{hyperref}
\usepackage{url}
\usepackage{booktabs}
\usepackage{multirow}
\usepackage{cleveref}
\usepackage{censor}
\usepackage{makecell}
\hypersetup{
  breaklinks=true,
  colorlinks=false
}
\usepackage{makecell}
\usepackage[ruled,vlined]{algorithm2e}
\usepackage{tablefootnote}
\usepackage[para,online,flushleft]{threeparttable}
\usepackage{xcolor}
\usepackage{soul}

\DeclareMathOperator*{\argmin}{arg\,min}
\def\BibTeX{{\rm B\kern-.05em{\sc i\kern-.025em b}\kern-.08em
    T\kern-.1667em\lower.7ex\hbox{E}\kern-.125emX}}

\begin{document}
\title{
Stein-based Optimization of Sampling Distributions in Model Predictive Path Integral Control
 {\footnotesize}
 \thanks{J. Aldrich and O. C. Jenkins are with the Laboratory for Perceptive RObotics and Grounded REasoning SystemS and Robotics Department at the University of Michigan, Ann Arbor, USA. This work was supported in part by Amazon, Ford Motor Company, and a National Science Foundation Graduate Research Fellowship.}
}

\author{Jace Aldrich \hspace{1cm} Odest Chadwicke Jenkins
}
\maketitle

\begin{abstract}

This paper introduces a method for Model Predictive Path Integral (MPPI) control that optimizes sample generation towards an optimal trajectory through Stein Variational Gradient Descent (SVGD). MPPI relies upon predictive rollout of trajectories sampled from a distribution of possible actions.  Traditionally, these action distributions are assumed to be unimodal and represented as Gaussian.
This assumption can lead to suboptimal rollout predictions due to sample deprivation and, in the case of differentiable simulation, sensitivity to noise in the cost gradients. 
By introducing SVGD updates in between MPPI environment steps, we present Stein-Optimized Predictive Path-Integral Inference (SOPPI), an MPPI/SVGD algorithm that can dynamically update noise distributions during rollouts to better capture action sampling distributions without an excessive increase in computational requirements. 
We demonstrate the efficacy of SOPPI through experiments on a planar cart-pole, 7-DOF robot arm, a planar bipedal walker, and a Unitree G1 29-DOF Robot.  Results indicate improved inference of an underlying multi-modal action distribution compared to other Stein-based MPPI algorithms, with improved system performance and similar or improved efficiency across a range of hyper-parameters.
\end{abstract}


\section{Introduction}

Model Predictive Control (MPC) methods, along with other optimal control methods, have become commonplace in several domains for robotics, ranging from simple automotives to even high-degree-of-freedom humanoid robots. The space of MPC-based methods includes an information theoretic nonparametric sampling method, Model Predictive Path Integral Control (MPPI) \cite{7487277}. While the performance of MPPI, both in resultant trajectories and computation time, is stellar, it is often only applied to relatively low-dimensional spaces or is abstracted to a lower dimension to make the problem more feasible.  We aspire to realize MPPI methods suitable for real-time robust inference for more complex robotic platforms, such as humanoid robots.

The key concept of MPPI involves information maximization over a projected set of rollout trajectories. 
From a current state estimate and initial action sequence, MPPI will heavily perturb the action sequence, generating a distribution of trajectories which are then rolled out via state dynamics. MPPI then aims to minimize the Kullback–Leibler (KL) Divergence \cite{williams2018information} (a measure of the difference between two distributions) between this rollout distribution and an optimal trajectory distribution, identified by a cost function. Functionally, this equates to performing an information maximization expectation to pick a trajectory equal to a weighted sum of the sampled trajectories. The perturbations, however, are generated from Gaussian distributions, which focus sampling primarily around the mean, which is not often representative of the true optimal distribution \cite{trevisan2024biased}. Particularly for constrained high degree-of-freedom non-linear systems, MPPI becomes susceptible to the curse of dimensionality, requiring exponentially more samples to be effective \cite{mohamed2025towards}. 
\begin{figure}[tb]
    \centering
    \includegraphics[width=0.95\linewidth]{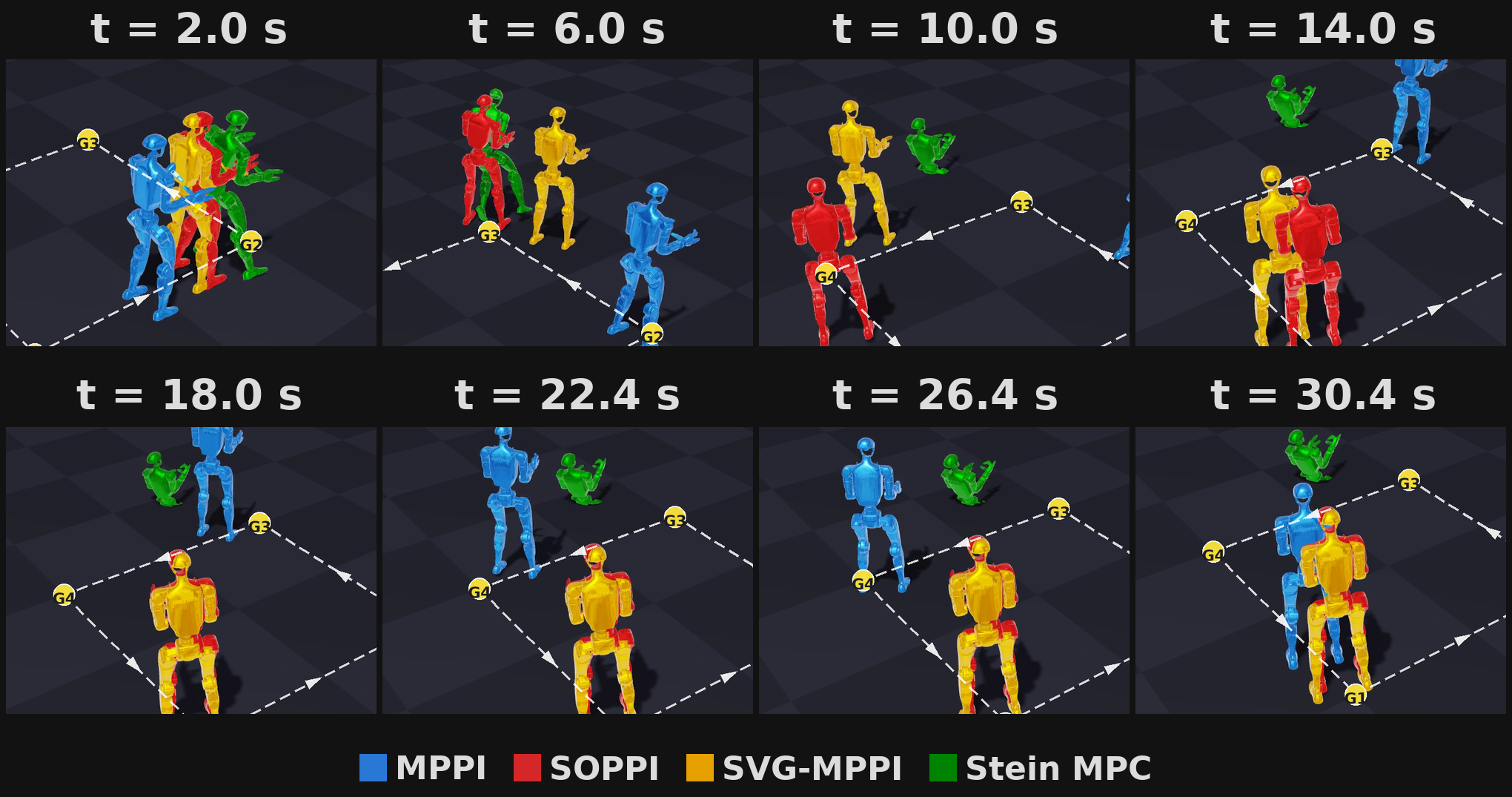}
    \caption{A sequence of a 2D Unitree G1 29-DOF walking in a square pattern. SOPPI converges before other algorithms (t=14.0s) followed by SVG-MPPI (18.0s), and eventually MPPI (after 30.4s). Stein-MPC frequently falls after 6.0s, but sometimes converges (approximately 10\% of the time)}
    \label{fig:walker-stairs}
\end{figure}

For similar problems relating to gradient based optimization Stein Variational Gradient Descent (SVGD), or Stein Variational Inference \cite{liu2016stein} has become increasingly popular, wherein as part of an optimization step, a kernel is used to force particles to spread out. This kernel ensures a more complete representation of a sampled space and prevents mode collapse, often improving particle efficiency. There are several implementations of SVGD within MPC, highlighted by \cite{lambert2020stein}, and a few within MPPI \cite{miura2024spline,honda2024stein}. 

However, these representations each have their own issues. For strict MPC implementations, the repulsive kernel force often loses magnitude quickly as it must differentiate particles over several sets of dimensions (the horizon, number of rollouts, and state and/or action dimensions) \cite{lambert2020stein}. These high dimensions and smaller gradients then make kernel distances much more arbitrary, limiting the effectiveness of the repulsive force in SVGD. Additionally, the Stein component of the gradient descent update must be recursively updated through the cost functions and dynamics over the entire horizon, adding significant computation time. 

Further, because SVGD requires gradients, the rollout simulation must be differentiable by some method, leaving it vulnerable to exploding or vanishing gradients \cite{xu2022accelerated}, not to mention the increased memory requirement to compute them. There have been attempts to merge SVGD with MPPI as well \cite{miura2024spline,honda2024stein}, however, for the aforementioned reasons, they perform SVGD on a small subset of particles, or omit most of the kernel terms, which can minimize, if not entirely eliminate, the benefits of SVGD.

In this paper, we present Stein-Optimized Predictive Path-Integral Inference (SOPPI), an algorithm that combines SVGD with MPPI, optimizing the action noise generation to create a more faithful action sample distribution compared to baseline MPPI. Compared to existing Stein-based methods, SOPPI performs SVGD updates during rollouts instead of after them, learning a new action distribution at each time-step. This process better preserves a multi-modal distribution and kernel meaning than other methods, and alleviates some concerns regarding exploding/vanishing gradients.

We evaluate the performance of SOPPI and state-of-the-art MPPI/MPC algorithms for a planar cart-pole system to study the effects of particle counts and sample efficiency between the algorithms.
Our findings from these tests suggest that there is a statistically significant improvement at the 95\% level in SOPPI's effectiveness over baseline MPPI and other methods, even when operated at lower particle counts. We also study higher degree-of-freedom (DOF) systems: a block pushing task with a 7-DOF robotic arm and a two-dimensional bipedal locomotion task with a 7-link, 6 DOF walking robot. We explore the effect of uncertainty in both: Gaussian noise added to gradients in the pushing task, and imperfect dynamics gradients in the locomotion task.
We lastly explore SOPPI on a Unitree G1 (29-DOF) in Nvidia Newton, a fully differentiable simulator, as an implementation optimized to evaluate computation time and efficiency.
As a whole, the experiments demonstrate that SOPPI more effectively covers the action space and generates improved control solutions, without significantly increasing computation time over other methods.

\begin{figure*}[!t]
    \centering
    \includegraphics[width=0.95\linewidth]{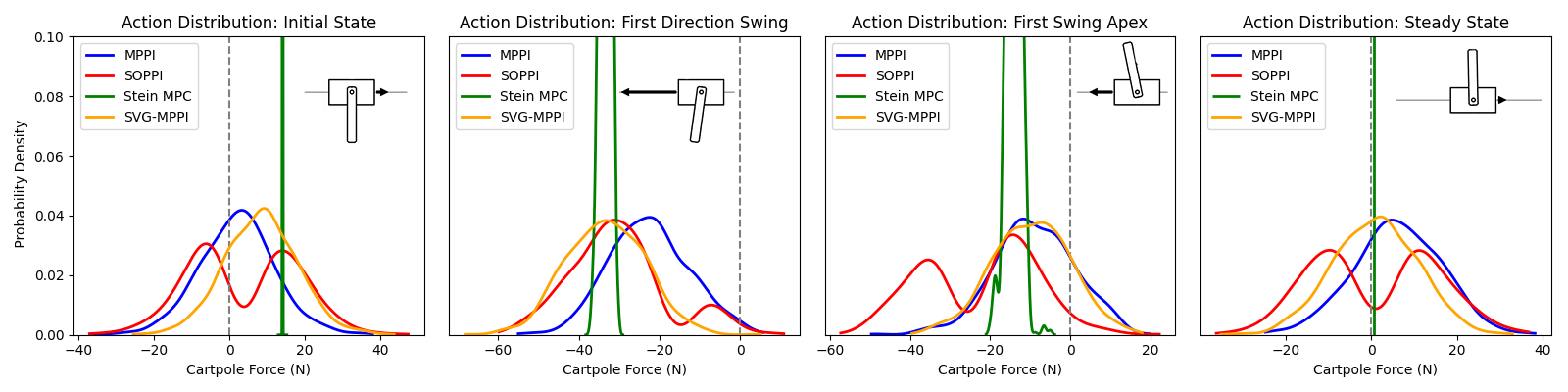}
    \caption{Inferred Action Distributions: Cart-pole action distribution at different points in the cart-pole's swing for all control algorithms evaluated in Section~\ref{section:cartpole_experiments}. An image of the cart-pole of each state is shown in the top right. Visualization code provided by \cite{mppi_repo}. All distributions are at similar states (not times) and are for the next planned step in the environment. SOPPI maintains multi-modality where needed, whereas other methods do not. 
    }
    \label{fig:action-noise}
\end{figure*}

\section{Related Work}
\subsection{MPC and MPPI}
Model Predictive Control (MPC) represents a collection of controller types that optimize trajectories via an explicit model \cite{holkar2010overview}. One of the most common MPC methods is under direct optimization, for example, a quadratic programming problem over a system's state space. These methods function on both linear and non-linear systems, although linear models are simpler and more consistent. \cite{morari1999model}.
MPPI, on the other hand, is a sampling-based approach for stochastic trajectory optimization, as introduced by Williams et al. in \cite{7487277}. MPPI algorithms generate several sample trajectories that are evaluated and weighted to minimize the cost function instead.
\subsection{MPPI-based Methods} 
Several versions of MPPI expand upon the original work \cite{7487277}. In particular, Tube MPPI \cite{williams2018robust} runs multiple controller layers to create more trajectory robustness guarantees. Constrained Covariance Steering MPPI \cite{9867864} builds upon this further, adding covariance tracking to Tube MPPI. Robust MPPI \cite{9349120} introduces additional safeties and disturbance rejection by importance sampling and augmented state spaces. Reinforcement Learning (RL) MPPI methods  \cite{qu2023rl,wang2024residual} attempt to improve sample efficiency though reinforcement learning by training an agent to generate trajectories, instead of baseline MPPI's normal distribution sampling. However, these methods suffer from normal RL drawbacks such as training data, time, and constraints. Unscented MPPI \cite{mohamed2025towards} implements an Unscented Transform to help manage uncertainty during rollouts, propagating mean and uncertainty throughout sampling. The aforementioned versions provide a reasonable representation of the space of MPPI methods at a high level, noting that these existing methods often assume the action distribution for sampling is Gaussian or unimodal.


\subsection{Variational Inference Methods}

On the other end of optimization lie gradient descent methods, in particular, stochastic variational inference \cite{hoffman2013stochastic}. This method operates in the probability space, and shifts a distribution towards an optimal one via gradient-based update rules and minimizing KL-divergence, a measure of the difference between two distributions. Stein Variational Gradient Descent (SVGD) \cite{liu2016stein} applies similar update rules to variational inference, but also adds kernel terms. These terms encourage particles to avoid mode collapse, increasing particle efficiency and assisting in multi-modal distributions.

Uniting SVGD and MPC is Stein MPC \cite{lambert2020stein}, which treats MPC as a Bayesian inference problem. This combination allows the application of SVGD instead of traditional gradient descent, which assists in complexity and multi-modality in trajectory selection. However, both represented in this work and the Stein MPC authors' own attempts \cite{lambert2020stein}, kernels across a high dimension, long horizon problem with a large number of samples quickly lose magnitude. The high dimensionality asks the kernel to cover too large a solution space, which reduces the overall coverage of the method. Hence, Stein MPC scales less effectively with respect to solution space coverage if the kernel is applied over the entire problem. 

Figure~\ref{fig:action-noise} shows an example of this effect for the cart-pole experiments in Section~\ref{section:cartpole_experiments}.  Stein MPC devolves into an unimodal action distribution, whereas SOPPI accurately enforces the true bimodality (swinging up on either side). This difference suggests that SOPPI is helping mitigate the kernel issue that hinders Stein MPC.
SVGD and Stein MPC have also been extended for trajectory optimization~\cite{aoyama2024second,pavlasek2023ready,power2024Svo} and belief propagation for multi-robot coordination~\cite{10465621}.


\section{Methods}
\subsection{MPPI}
In this paper, we build upon the original derivation of MPPI in \cite{7487277}. We consider the stochastic optimal control problem for the general, discrete-time dynamic system operating in a continuous space
\begin{equation*}
    x_{t+1}=F(x_t,v_t)
\end{equation*}
where $x_t,v_t$ represent system state and controls at time $t$ and $F$ denotes the system dynamics. Additionally, $v_t$ is assumed to be normally distributed $v_t\sim N(u_t,\sigma^2)$ where $u_t$ represents a targeted input, and $\sigma^2$ represents some process noise. We then aim to optimize some control trajectory $U=(u_0,u_1,...,u_{T-1})$ of length $T$ via
\begin{equation*}
U^*=\argmin_{U\in\mathbb{U}}\mathbb{E}\left[\phi(x_T)+\sum_{t=0}^{T-1}\mathcal{L}(x_t,v_t)\right]
\end{equation*}
where $\mathcal{L}(x_t,u_t)$ and $\phi(x_T)$ represent running and terminal cost functions respectively, defined as
\begin{equation}\label{eq:cost-run}
    \mathcal{L}(x_t,u_t)=x_t^TQx_t+v_t^TRv_t
\end{equation}
\begin{equation}\label{eq:cost-term}
    \phi(x_T)=x_T^TQ_Tx_T
\end{equation}
where $R, Q, Q_T$ represent weight matrices for action costs, running state costs, and terminal state costs. We can then define $S(\tau)$, a cost-to-go function, as dependent on an entire trajectory such that
\begin{equation}\label{eq:cost-to-go}
    S(\tau)=\phi(x_T)+\sum_{t=0}^{T-1}\mathcal{L}(x_t,v_t)
\end{equation}
where $\tau$ represents a trajectory in the form
\begin{equation*}
    \tau=(x_0,v_0,x_1,v_1,...,x_T)
\end{equation*}
To optimize, we first must generate a set of $K$ sample trajectories to minimize KL-Divergence towards some optimal trajectory.
This process is done by sampling a zero-mean Gaussian distribution $\varepsilon$ with some variance $\sigma^2$, and adding it to some initial guess of a trajectory $u$. These trajectories are rolled out and have their costs computed. Then, the original MPPI derivation yields that for these sample trajectories, each with its own cost-to-go, we can approximate $u^*$ via an iterative update for $K$ samples
\begin{equation}\label{eq:u-update}
    u^*=u+\sum_{k=1}^Kw_k\varepsilon^k
\end{equation}
where $w_k$ represents a weight for the $k$-th trajectory with cost $S_k$, solved as
\begin{equation}\label{eq:weights}
    w_k=\frac{\exp(-\frac{1}{\lambda}(S_k-\beta))}{\sum_{j=1}^k\exp(-\frac{1}{\lambda}(S_j-\beta))}
\end{equation}
where $\beta=\min_{k\in \mathbb{K}}(S_k)$. Note that there are several MPPI variations that apply filtering to the update step as well, which in some cases may improve system performance.
\subsection{SVGD}
Similar to MPPI, SVGD attempts to optimize some particle set by minimizing KL-Divergence. For a set of particles $\{\theta^i\}_{i=1}^k$ it follows the iterative update rule of
\begin{equation*}
    \theta^i\leftarrow\theta^i+\epsilon\phi^*(\theta^i)
\end{equation*}
where $\epsilon$ represents a step size, and $\phi^*(\cdot)$ defines the optimal perturbation to reduce KL-divergence via kernel functions, and can be approximated for a set of $K$ particles via
\begin{equation}\label{eq:SVGD}
    \hat\phi^*(\theta)=\frac{1}{K}\sum_{j=1}^K\left[ k(\theta^j,\theta)\nabla_{\theta^j}\log p(\theta^j||x)+\nabla_{\theta^j}k(\theta^j,\theta) \right]
\end{equation}
where $k(\cdot)$ represents a valid kernel function. The first term is a scaled gradient log-likelihood of the particle's posterior, which drives the particles towards an optimal state. The second term is a repulsive force that prevents mode collapse and allows greater sample coverage. A full derivation can be found at \cite{liu2016stein}, with a high-level explanation at \cite{lambert2020stein}.


\subsection{SOPPI} 
Similar to \cite{miura2024spline}, the SOPPI Algorithm (Algorithm~\ref{alg:mppi}) we propose performs SVGD updates to improve the random samples generated for MPPI. Compared to methods like Stein MPC \cite{lambert2020stein}, SOPPI is designed to optimize the noise added to the distribution per time-step, not the entire trajectory itself. This choice results in simpler, shorter computations as gradients do not have to propagate through the entire horizon, reducing the issue of exploding or vanishing gradients. 

With modern automatic differentiation (e.g., Pytorch \cite{Ansel_PyTorch_2_Faster_2024}), analytical dynamics are simple to compute, but can become difficult to model with complex systems. Simulation tools such as MuJoCo \cite{todorov2012mujoco} and Gymnasium \cite{towers2024gymnasium} are widely used to execute dynamics, but are often non-differentiable, requiring solutions such as Recurrent Neural Networks (RNN) \cite{mukhopadhyay2019model} to approximate dynamics, usually with error and a high training cost. Differentiable simulators exist (e.g., Nvidia's Newton \cite{The_Newton_Contributors_Newton_GPU-accelerated_physics_2025}), but they often simplify operations or run below real-time \cite{newbury2024review}, and generally result in a more difficult to achieve implementation. However, by limiting the gradients to a single time-step rather than the entire horizon, SOPPI is less susceptible to the noisy or imprecise gradients of these differentiation methods, as demonstrated by results in Sections \ref{sec:arm} and \ref{sec:2d_walker}.


This approach also forces the SVGD kernel to measure distance per time-step, instead of over the entire problem. Whereas Stein MPC's kernel function can become arbitrary and fail to disperse particles, SOPPI's does not, representing multi-modality much better, as depicted in Figure \ref{fig:action-noise}. By envisioning the SVGD computation as a series of small sub-problems at each time-step, we can ensure that the kernel operation appropriately prevents mode collapse. Thus, this method can improve sampling efficiency and better represent candidate solutions compared to a standard normal distribution sample.

SOPPI uses the same cost structure as baseline MPPI as seen in Equations \ref{eq:cost-run} and \ref{eq:cost-term}. Similar to Miura et al.~\cite{miura2024spline}, we apply SVGD updates to the trajectories; however, in our case, we apply SVGD updates at each time-step during the rollout, not after. To start, SOPPI performs  MPPI for a single time-step in the environment to gain a set of samples. 





With these samples, we can derive a cost-likelihood function $\mathcal{L}_s$ for SOPPI from the MPPI costs to represent our trajectory's distribution, such that
\begin{equation}
    \mathcal{L}_s(x_t,v_t)=\exp(-\alpha\mathcal{L}(x_t,v_t))
\end{equation}
where $\mathcal{L}$ is defined in Equation \ref{eq:cost-run}. Thus, the gradient of the log-likelihood as listed in Equation \ref{eq:SVGD} is the negative scaled gradient of our running cost function.
Additionally, we use a standard radial-basis (RBF) kernel parameterized by $\sigma$:
\begin{equation}
    k(v_t^j,v_t)=\exp\left(-\frac{||v_t^j-v_t||}{2\sigma^2}\right)
\end{equation}

From these equations, we compute the SVGD update for SOPPI defined in \ref{eq:SVGD}, and apply it to the sampled action particles immediately. Then, we iteratively re-compute the initial environment step with the new actions, before moving to the next environment step. This process is repeated throughout some horizon, after which the weighting defined in equations \ref{eq:u-update} and \ref{eq:weights} are used to select an optimal trajectory. This process effectively creates sequential sub-problems with a lower search dimension than if the entire trajectory were searched at once.
The entire algorithm is depicted in Algorithm \ref{alg:mppi}.
\SetAlgoHangIndent{3.2cm}

\begin{algorithm}[!htb]

\KwData{
\begin{tabular}{@{}l@{\hspace{1em}}l@{}}
$K$: \# samples & $f(x,u)$: Dynamics \\
$N$: \# time-steps & $k(v,v)$: Kernel \\
$M$: \# SVGD updates & $\sigma^2$: Sampling variance \\
$U_{init}$: Init. control seq. & $\lambda$: Temperature \\
\end{tabular}
}
%
\While{task not completed}{
    $\epsilon \sim \mathcal{N}(0, \sigma^2)$\;
    $v\gets U_{init}+\epsilon$\;
    \For{$k \gets 0$ \KwTo $K-1$ $\text{in parallel}$}{
        $x^k \gets x_0$\;
        $S^k \gets 0$\;
        \For{$t \gets 0$ \KwTo $N-1$}{
            \For{$i \gets 0$ \KwTo $M-1$}{
                $x_{t+1}^k \gets f(x_t^k,v_t^k)$\;
                $S_s^{k} \gets\mathcal{L}(x_t^{k+1},v_t^k)$\;
                $\hat{\phi}^*(v)\gets\frac{1}{K}\sum_{j=1}^K\big[-k(v^j,v)\nabla_{v^k_t}S_s^k+\nabla_{v^j}k(v^j,v) \big]$\;
                $v_t^k \gets v_t^k+\epsilon\hat{\phi}(v) $\;
            }
            $x_{t+1}^k \gets f(x_t^k,v_t^k)$\;
            $S^{k} \gets S^k + \mathcal{L}(x_t^{k+1},v_t^k)$\;
        }
        $S^k \gets S^k + \phi(x_N^k)$\;
    }
    $\beta \gets\min_{k\in \mathbb{K}}(S_k)$\;
    $w_k\gets\frac{\exp(-\frac{1}{\lambda}(S_k-\beta))}{\sum_{j=1}^k\exp(-\frac{1}{\lambda}(S_j-\beta))}$\;
    $u^*\gets u+\sum_{k=1}^Kw_k\varepsilon^k$\;
    $x_0\gets f(x_0,u^*_0)$\;
    $U_{init} \gets \begin{bmatrix}
        u_{1:N}^* & 0
    \end{bmatrix}$\;
}
\caption{SOPPI}\label{alg:mppi}

\end{algorithm}

\section{Experiments}
We conducted a series of simulation experiments comparing SOPPI to base MPPI \cite{7487277}, SVG-MPPI \cite{honda2024stein} and Stein MPC \cite{lambert2020stein}. We evaluate multiple tasks using a range of gradient computation methods: a single DOF cart-pole upswing (analytical gradients), 7-DOF robot arm box pusher (RNN), a planar 6-DOF bipedal walker (RNN), and a 29-DOF Unitree G1 (Nvidia Newton).
All simulations were conducted through Pytorch \cite{Ansel_PyTorch_2_Faster_2024} and analytical dynamics or the Gymnasium \cite{towers2024gymnasium} wrap of MuJoCo \cite{todorov2012mujoco} with a trained recurrent neural network. All tests were performed with an i7-13700K CPU, accelerated by a Nvidia A6000 GPU.

\subsection{Cart Pole} \label{section:cartpole_experiments}
For the cart-pole swing-up task, we used an analytical dynamics model \cite{florian2007correct} with an upright pole angle of $\theta=0$, and the starting, downward pole angle as $\theta=\pi$. The cart's lateral ($x$) position starts and ends at $x=0$. Visualizations of the cart-pole system are depicted in Figure \ref{fig:action-noise}.

\begin{table*}[!tb]
    \centering
    \begin{threeparttable}
    \caption{Cart Inverted Pendulum System Response for a Horizon (H) of 80 steps or 1.6 seconds, and various particle counts (K). Statistically significant (95\%) best results are bolded.}
    \label{tab:single-cart-results}
    \begin{tabular}{|c||c|c|c|c|c|c|}
    \hline
    Trial & MSE$(x)$ & $t_{s,x,0.25m}$ & $t_{s,x,0.5m}$& MSE$(\theta)$ & $t_{s,\theta,5\%}$ & $t_{s,\theta,10\%}$ \\
    \hline\hline
    MPPI H=80, K=500 & 0.58 & 6.77 $\pm$ 0.94$^\dagger$ & 4.33 $\pm$ 0.02 & 1.36 & 2.68 $\pm$ 0.02 & 1.97 $\pm$ 0.02 \\
 SOPPI H=80, K=500 & \textbf{0.53} & 5.20 $\pm$ 0.80$^\dagger$ & 3.46 $\pm$ 0.73 & \textbf{1.29} & \textbf{2.65 $\pm$ 0.02} &\textbf{1.87 $\pm$ 0.04} \\
    Stein MPC H=80, K=500 & 1.26 & 5.53 $\pm$ 0.13 & 3.76 $\pm$ 0.13 & 1.72 & 4.22 $\pm$ 0.14 & 3.90 $\pm$ 0.02 \\
    SVG-MPPI H=80, K=500 & 0.78 & 4.34 $\pm$ 0.18 & 2.95 $\pm$ 0.03 & 1.40 & 2.99 $\pm$ 0.09 & 2.89 $\pm$ 0.03 \\
    \hline\hline
    MPPI H=80, K=1,000 & 0.56 & 6.64 $\pm$ 0.69 & 4.03 $\pm$ 0.68 & 1.36 & 2.69 $\pm$ 0.01 & 1.98 $\pm$ 0.02 \\
 SOPPI H=80, K=1,000 & \textbf{0.53} & 5.66 $\pm$ 0.51 & 3.93 $\pm$ 0.66 & \textbf{1.30} & \textbf{2.65 $\pm$ 0.01} &\textbf{1.87 $\pm$ 0.01} \\
    Stein MPC H=80, K=1000 & 1.32 & 5.56 $\pm$ 0.06 & 3.88 $\pm$ 0.13 & 1.73 & 4.07 $\pm$ 0.12 & 3.90 $\pm$ 0.02 \\
    SVG-MPPI H=80, K=1000 & 0.79 & \textbf{4.28 $\pm$ 0.27} & \textbf{2.97 $\pm$ 0.05} & 1.41 & 2.97 $\pm$ 0.04 & 2.89 $\pm$ 0.02 \\
    \hline
    \end{tabular}
    \begin{tablenotes}
    Note: $t_{s,x}$ and $t_{s,\theta}$ represent settling times of $x$ and $\theta$, followed by criteria: a meter range for $x$ or percentage of the step range ($\pi$) for $\theta$.
    Metrics with a $\dagger$ did not converge on two trials.
    \end{tablenotes}
    \end{threeparttable}
\end{table*}

We performed a range of tests on all four algorithms focusing on varying sample count and horizon time-steps. We present results at a horizon of 80 time-steps or 1.6 seconds, and at 500 and 1000 particles to demonstrate differences in particle efficiency between the algorithms. Additionally, for the gradient based algorithms, we present results with a learning rate of $0.05$ over $100$ iterations. All remaining hyper-parameters and costs were identical across tests. Numerical results from these tests can be noted in Table \ref{tab:single-cart-results} depicting settling times $t_s$ for both $x$ and $\theta$ for multiple criteria (for example, 0.25 m on $x$ or $5\%$ of the $\theta$ range). 

From the results in Table \ref{tab:single-cart-results}, we note that SOPPI performs statistically better (Welch's t-test) with respect to the cart pole's state mean-squared errors and angular settling times at identical particle counts. This metric indicates an improved overall error tracking by SOPPI relative to the compared algorithms. Figure \ref{fig:action-noise} provides further insight, highlighting SOPPI's multi-modal action distribution. SOPPI matches a mode to the other Stein methods when appropriate (e.g., during the swing), or splits the modes when those actions are inappropriate (e.g., initial/steady state) to accelerate a swingup or better maintain equilibrium.


For the x position settling time, SOPPI converges faster than all but SVG-MPPI, but SOPPI still outperforms it on overall x tracking error. We attribute this discrepancy to the single-modal nature of SVG-MPPI. The action distribution as depicted in Figure \ref{fig:action-noise} shows a propensity of SVG-MPPI to excessively favor a single mode, balancing the x position faster, but at the cost of every other metric.
 
Lastly, these trends hold when SOPPI operates at less particles. SOPPI at 500 particles performs statistically better than other methods at 1,000 particles on x and $\theta$ MSE and the angle settling times, with the exception of MPPI's 10\% angle settling time (numerically better only). Thus, we conclude that SOPPI's multi-modal action distribution covers a wider solution space, increasing the algorithm's particle efficiency.

\subsection{Arm Pushing Task}\label{sec:arm}
To increase the complexity of the system over the cart-pole, we elected to test the algorithm on a simulated planar pushing task on a block with a Franka Panda arm based on Berenson et al.~\cite{Berenson_Van_der_Merwe_Huang}. Since this system is the most stable of those tested, we also elected to test uncertainty handling via trials with noise injected into the gradients.


To simplify the dynamics at runtime while maintaining gradients, we trained a recurrent neural network \cite{mukhopadhyay2019model} to approximate the next state from a MuJoCo simulation, given a current state and action. In this formulation, states represent the block's pose $\mathbf x = \begin{bmatrix} x & y & \theta\end{bmatrix}^\top\in \text{SE}(2)$ and actions $\mathbf u = \begin{bmatrix} p & \phi & \ell\end{bmatrix}^\top\in \mathbb R^3$ are represented by $p$ corresponding to the lower edge pushing location of the block, $\phi$ corresponding to the pushing angle, and $\ell$ corresponding to the pushing length as a fraction of the maximum allowed length (0.1 m in this case). In all tests, the arm was tasked with pushing the block from the pose $[0.4,0,\frac{\pi}{5}]$ to $[0.8,0,0]$.

\begin{figure}[!htb]
    \centering
    \includegraphics[width=0.95\linewidth]{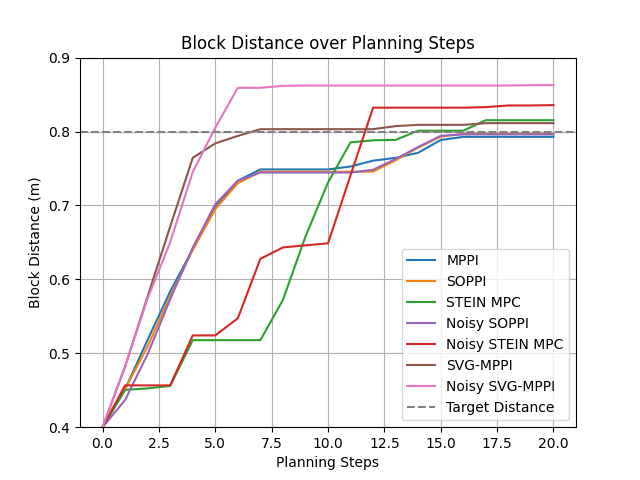}
    \caption{Median trial pushing distance error compared between algorithms}
    \label{fig:pushing-chart}
\end{figure}

Since this task was inherently more stable than the cart-pole task, we elected for a shorter horizon and higher end sample count, 10 time steps (0.2 seconds) and 1000 samples, for both MPPI and SOPPI. 
All other hyper-parameters and cost functions were identical between trials for all 4 algorithms. Table \ref{tab:pushing-tab} and Figure \ref{fig:pushing-chart} 
depict numeric results.


\begin{table}[!htb]
    \centering
    \begin{threeparttable}
     \caption{Steady State Distance Errors for the pushing task}
    \label{tab:pushing-tab}
    \begin{tabular}{|c|c|c|}
    \hline Algorithm& \makecell{Mean End Distance \\ Error (mm)}& \makecell{Standard Deviation \\  of End Distance (mm)}\\\hline\hline
 MPPI& 8.02&5.15\\\hline
 Stein MPC& 19.36&16.06\\\hline
 Noisy Stein MPC& 37.43&20.94\\\hline
 SVG-MPPI& 12.08&5.61\\\hline
 Noisy SVG-MPPI& 62.12&8.47\\\hline
 SOPPI (ours)& \textbf{4.48}&4.96\\\hline
 Noisy SOPPI (ours) & 5.26&4.31\\ \hline\end{tabular}
   \end{threeparttable}
\end{table}

Overall, from Table \ref{tab:pushing-tab} we note that the mean steady-state response errors on the pushing direction with SOPPI are slightly better than MPPI and much better than any other method, with or without noise. Errors in the lateral direction and desired angle are quite similar across all systems, with the exception of a larger lateral error in MPPI. From Figure \ref{fig:pushing-chart}, SOPPI reaches steady state slightly faster than MPPI, but compared to other methods, SOPPI and MPPI do not overshoot. Given the reach of the arm, an overshoot is unrecoverable, as the arm physically cannot reach the other side of the block. Hence, SOPPI and MPPI act more cautiously than the other algorithms, especially in the case of added noise. Stein MPC and SVG-MPPI both overshoot significantly with noise added to the gradients, whereas SOPPI has little change in performance. Combined with the cart-pole results, we can empirically note that SOPPI better handles uncertain gradients than some other direct optimization methods. Simulations often simplify dynamics, and thus their gradients, compared to the real-world \cite{newbury2024review}, so successful performance under uncertainty is quite important for operation on real systems.
Experiments for the 2D walker further highlight successful operation under gradient noise.


\subsection{2D Walker Task}\label{sec:2d_walker}
We implement the same algorithms on the Walker2D gymnasium environment \cite{towers2024gymnasium}, which is commonly used in the reinforcement-learning community. For this system, however, we apply control directly at the joint torque level. Due to the unstable dynamics of the system's 6 joints though, we implement a cost structure similar to \cite{alvarez2025real}, wherein we apply a reference input trajectory around some known gait. Since this system is well studied in the RL community, we trained a Proximal Policy Optimization (PPO) \cite{DBLP:journals/corr/SchulmanWDRK17} model to generate reference gait, from the Stable Baselines3 package\cite{stable-baselines3}. Additionally, our cost structure was modified from \ref{eq:cost-run} to

\begin{equation*}
    \mathcal{L}(x_t,u_t)=x_t^TQx_t+(v_t-u_{ref})^TR(v_t-u_{ref})
\end{equation*}
where $u_{ref}$ was computed by the PPO model, rolling out from the current state with its own generated actions. To prevent collapsing the distribution on the reference input, we ensure that $Q$ has a much higher magnitude than $R$ so that states weight the cost function more than the inputs. During sampling, we also modified the initial input for the last time-step to be the output of the PPO model at the next to last time-step's states. So the last line of Algorithm \ref{alg:mppi} becomes

\begin{equation*}
    U_{init} \gets \begin{bmatrix}
        u_{1:N}^* & \text{PPO}(x_{N-1})
    \end{bmatrix}
\end{equation*}

For gradients, we trained another RNN to approximate the dynamics of the system for a single step (the duration of gradients required). However, we used MuJoCo simulations for rollouts, as given the chaotic nature of the walker robot, we found MuJoco to be more accurate over a long-horizon. This issue could be reduced with more training and data, but was not required for SOPPI to function. Regardless, there will always be some degree of error in a learned dynamics system, and gradients usually explode or vanish over a long enough horizon \cite{xu2022accelerated}, so these issues will exist in some capacity. Managing these simulated gradients is an active field of research, particularly with differentiable simulation \cite{newbury2024review}. SOPPI is one such approach to apply control in spite of these issues, as evidenced by its performance.

\begin{figure}[!t]
    \centering
    \includegraphics[width=1.0\linewidth]{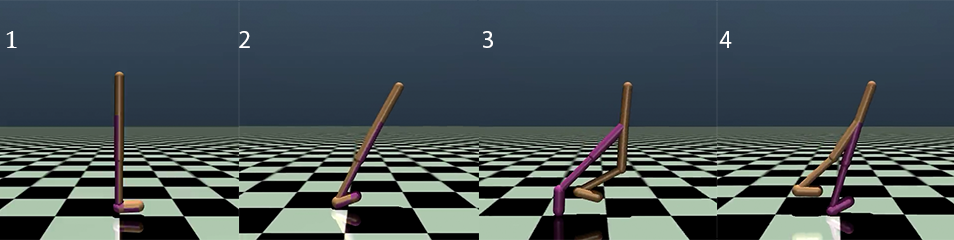}
    \caption{Sample 2D walker gait from initial position. 
    }
    \label{fig:walker-visual}
\end{figure}

Lastly, we tuned our cost matrix, $Q$ to prioritize maximizing the height of the walker's torso, and added forward kinematic costs which incentivized the robot's feet to be further apart. Both modifications led to more stable, continued steps. We conducted eight trials for each algorithm with all hyper-parameters and cost functions equal. We elected for a 50 time-step horizon and 1,000 samples for each case. The results of these trials are depicted in Table \ref{tab:pushing-tab}, with visuals of the system walking depicted in Figure \ref{fig:walker-visual}. We consider the "walking time" metric to denote the duration for which the walker remains upright, and consider the end of walking time as the last peak before the torso height falls below 1 meter. 

\begin{table}[!]
    \centering
    \begin{threeparttable}
    \caption{Walker Task Stable Gait Time. Longer times indicate further distance traveled upright}
    \label{tab:walker-tab}
    \begin{tabular}{|c|c|c|c|}
    \hline
    Algorithm&  \makecell{Mean \\ Walking Time}&\makecell{Standard \\Deviation} &\makecell{Median \\ Walking Time}\\ \hline \hline 
         MPPI& 
     20.37 s&13.33 s&15.20 s\\ \hline 
 Stein MPC& 3.15 s&3.02 s&1.50 s\\ \hline 
 SVG-MPPI& 9.81 s&15.51 s&3.41 s\\ \hline
 SOPPI& \textbf{44.41} s& 24.83 s&\textbf{39.49 s}\\\hline 
 \end{tabular}
    \end{threeparttable}
\end{table}


In our experiments, we observe that on average, SOPPI trials were able to walk longer than trials of the other algorithms tested, although not indefinitely. Typically, a trial failed (denoted by the walker falling to the ground) by either a slip or stall in momentum, which we believe to be caused by failure to sample a valid trajectory. Thus, we believe that with the improved sampling of SOPPI, the walker is overall more stable and better able to maintain its upright position.

The Stein MPC and SVG-MPPI algorithms, however, are not at all successful. They try to optimize over an entire horizon (x in this case), but given the chaotic nature of the system and accumulated errors on the learned dynamics, cannot optimize successfully. However, in SOPPI's case, the single step optimization improves upon MPPI where the other methods could not, leading to much better results.

A one-tailed Welch's t-test on the walking times of the different algorithms compared to SOPPI yields p-values of 0.023, 0.0016,  and 0.0047, for MPPI, Stein MPC, and SVG-MPPI respectively, indicating significant performance improvements at the 95\% level for SOPPI. We do, however, notice a large variance in the results for all algorithms, which we expect is due to the chaotic, unstable nature of the system. Similarly to how neural networks were only used for gradients when needed for stability, we expect that the sampling requirements for perfect stability are higher than conducted in our trials. In addition, many real-world systems have more rigorous reference gaits than our learned one (a "skipping" motion as seen in image 3 of Figure \ref{fig:walker-visual}). For this reason, we expect that implementing SOPPI on systems designed for the real-world (whether in simulation or not), which have the benefit of tested, stable gaits, would reduce this variance. Instead, especially with any remaining noise, SOPPI would lead to more optimized gaits, similar to the other systems presented in this paper. This trend is observed in practice by our last test case, a Unitree G1 Simulation.


\subsection{Unitree G1 (29-DOF) Walking}
We lastly implemented SOPPI on a 29-DOF Unitree G1 Robot in Nvidia Newton \cite{The_Newton_Contributors_Newton_GPU-accelerated_physics_2025} for three separate walking tasks. To avoid sampling over the full 29-DOF, we run SOPPI over the Unitree's Center of Mass (COM) trajectory, commanding the robot's forward, lateral, and rotational velocities with respect to a position goal. These commands are then passed to an underlying learned policy, provided by default of Nvidia Newton's Robot Policy Example \cite{The_Newton_Contributors_Newton_GPU-accelerated_physics_2025}.

The three tasks included a forward test (walking forward 3 meters), a slalom task (walking forward 2 meters at a time, alternating 1 meter from neutral right then left), and a square task (walking in a 2 meter by 2 meter square). For each task, we follow a similar cost structure as Equations \ref{eq:cost-run} to \ref{eq:cost-to-go}, following a goal COM position, shifting to the next goal when reaching the current one. However, we also add a stability cost to the height of the Unitree, z, of $\text{ReLU}(0.3-z)$, adding a linear cost if the COM falls below 0.3 meters.

\begin{figure}
    \centering
    \includegraphics[width=0.95\linewidth]{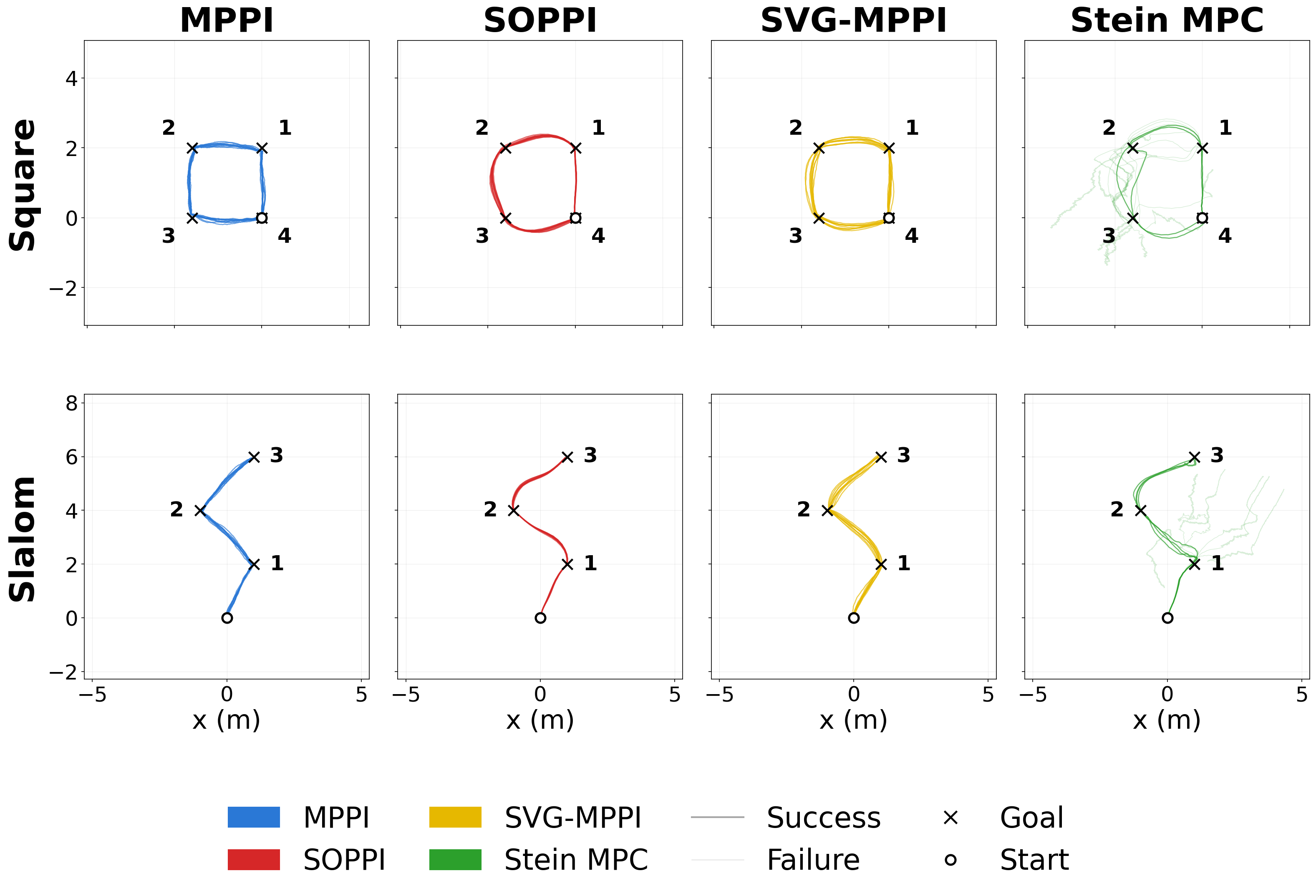}
    \caption{Trajectories for the Square and Slalom tasks on the Unitree G1 humanoid robot. Numbered x's mark waypoint goals in ascending order.}
    \label{fig:square_slalom}
\end{figure}

We tuned our weights to prioritize costs in the following order: stability, x-y COM position, COM angle, actuation cost. Table \ref{tab:unitree_table} shows numerical results for all trials, while Figure \ref{fig:square_slalom} shows a COM trace for the slalom and square tasks. Figure \ref{fig:walker-visual} highlights visual frames from a sample trial. Given the more stable underlying gait, we elected for a shorter horizon of 3 planning steps (with 4 execution only sub steps each). All tested environments had a clear linear path between goals, meaning a longer horizon would only significantly assist during switching goals or reaching one. Despite this shorter horizon, SOPPI experimentally demonstrates the ability to plan around these goal switches much more efficiently than other tested algorithms.

\begin{table*}[t]
\centering
\caption{
Controller comparison on Unitree G1 29-DOF (10 trials each). Mean$\pm$std over successful trials. \textbf{Bold} = best among controllers with 10/10 success. \textit{Italic} = Stein MPC metrics, computed over successful trials only.}
\label{tab:unitree_table}
\setlength{\tabcolsep}{4pt}
\begin{tabular}{ll|c|ccc|cc|cc}
\toprule
\textbf{Preset} & \textbf{Controller} & \textbf{Succ}
  & \makecell{\textbf{Cost of}\\\textbf{Transport}}
  & \makecell{\textbf{Execution}\\\textbf{Time (s)}}
  & \makecell{\textbf{Path}\\\textbf{Length (m)}}
  & \makecell{\textbf{Cross-Track}\\\textbf{Distance (m)}}
  & \makecell{\textbf{Speed}\\\textbf{(m/s)}}
  & \makecell{\textbf{Control Clock}\\\textbf{Time (ms)}}
  & \makecell{\textbf{Total Path}\\\textbf{Energy (J)}} \\
\midrule
  \multirow{4}{*}{\textbf{Forward}}
   & MPPI      & 10/10              & $2.051\pm0.043$          & $8.7\pm0.5$          & $3.0\pm0.0$             & $0.026\pm0.007$          & $0.34\pm0.02$          & $\mathbf{44.5\pm0.5}$   & $2167\pm54$          \\
   & SOPPI     & 10/10              & $\mathbf{1.695\pm0.049}$ & $\mathbf{3.7\pm0.0}$ & $\mathbf{2.9\pm0.0}$    & $\mathbf{0.014\pm0.003}$ & $\mathbf{0.78\pm0.01}$ & $48.8\pm0.5$            & $\mathbf{1758\pm50}$ \\
   & SVG-MPPI  & 10/10              & $1.980\pm0.061$          & $4.1\pm0.1$          & $\mathbf{2.9\pm0.0}$    & $0.037\pm0.014$          & $0.71\pm0.02$          & $45.0\pm0.6$            & $2061\pm63$          \\
   & Stein MPC & $5/10$   & $\textit{2.296}\pm\textit{0.219}$ & $\textit{3.6}\pm\textit{0.2}$ & $\textit{3.0}\pm\textit{0.0}$ & $\textit{0.015}\pm\textit{0.004}$ & $\textit{0.82}\pm\textit{0.04}$ & $\textit{131.0}\pm\textit{1.8}$ & $\textit{2416}\pm\textit{247}$ \\
\midrule
  \multirow{4}{*}{\textbf{Square}}
   & MPPI      & 10/10              & $2.329\pm0.027$          & $30.5\pm0.8$         & $\mathbf{8.1\pm0.1}$    & $\mathbf{0.048\pm0.010}$ & $0.26\pm0.01$          & $43.8\pm0.4$            & $6674\pm97$          \\
   & SOPPI     & 10/10              & $\mathbf{1.816\pm0.021}$ & $\mathbf{10.2\pm0.0}$ & $8.5\pm0.0$            & $0.165\pm0.005$          & $\mathbf{0.83\pm0.00}$ & $46.9\pm0.3$            & $\mathbf{5463\pm73}$ \\
   & SVG-MPPI  & 10/10              & $2.062\pm0.045$          & $13.2\pm0.3$         & $8.3\pm0.1$             & $0.121\pm0.019$          & $0.63\pm0.01$          & $\mathbf{43.8\pm0.6}$   & $6042\pm132$         \\
   & Stein MPC & $1/10$   & $\textit{11.210}$ & $\textit{29.0}$ & $\textit{13.7}$ & $\textit{0.246}$ & $\textit{0.47}$ & $\textit{127.7}$ & $\textit{54636}$ \\
\midrule
  \multirow{4}{*}{\textbf{Slalom}}
   & MPPI      & 10/10              & $2.126\pm0.038$          & $25.8\pm0.5$         & $\mathbf{7.9\pm0.0}$    & $\mathbf{0.055\pm0.010}$ & $0.31\pm0.01$          & $\mathbf{44.0\pm0.4}$   & $5966\pm109$         \\
   & SOPPI     & 10/10              & $\mathbf{1.820\pm0.030}$ & $\mathbf{10.0\pm0.1}$ & $8.2\pm0.0$            & $0.140\pm0.005$          & $\mathbf{0.82\pm0.00}$ & $47.4\pm0.3$            & $\mathbf{5269\pm79}$ \\
   & SVG-MPPI  & 10/10              & $1.987\pm0.062$          & $11.9\pm0.2$         & $8.0\pm0.1$             & $0.123\pm0.026$          & $0.67\pm0.01$          & $44.1\pm0.5$            & $5626\pm156$         \\
   & Stein MPC & $1/10$   & $\textit{2.320}$ & $\textit{9.2}$ & $\textit{8.7}$ & $\textit{0.216}$ & $\textit{0.94}$ & $\textit{126.9}$ & $\textit{7151}$ \\
\bottomrule
\end{tabular}
\\[4pt]
\end{table*}

Overall, we see similar trends as on other tests. SOPPI outperforms other algorithms highlighting a combination of performance and consistency. Stein MPC provides close performance, and sometimes a faster speed, but at the cost of several failure cases, succeeding half or less of the time. Of the successful controllers, SOPPI is a clear winner, having significantly lower energy usage and cost of transport, the commonly used metric to compare efficiency, defined as: $\text{CoT} =(\text{energy})/(\text{weight} \times \text{path length})$ \cite{collins2005efficient}. It also has significantly faster speeds and task execution times, all while maintaining a relatively similar control clock time (time to compute a control loop or next action). As expected, SOPPI takes longer to compute than SVG-MPPI or MPPI due to the extra computation, but not to an unreasonable level. 

SOPPI also has typically longer path lengths and cross-track distances (average perpendicular distance from a straight line between goals), but we attribute this result more to the tradeoff for success on other metrics. SOPPI's control output has the Unitree round the corners more heavily, which results in a longer path and more deviation from the shortest possible one, but with the benefit of executing that path significantly faster.
Effectively, SOPPI maintains momentum better between goals; a trend showcased in Figure \ref{fig:square_slalom}.

\section{Conclusions and Future Works}
In this paper, we presented SOPPI, a novel method that interweaves SVGD updates into MPPI rollouts, combining the benefits of SVGD's explicit optimization and MPPI's general computational simplicity/efficiency. 
We performed experiments on four systems, all demonstrating SOPPI's efficacy compared to MPPI \cite{7487277}, Stein MPC \cite{lambert2020stein}, and SVG-MPPI \cite{honda2024stein}. 
Specifically, the SOPPI's SVGD updates optimized the noise generation in MPPI algorithms to generate an improved performance, even in some cases where SOPPI operated at lower particle counts. SOPPI was also able to handle environments with noise and instability better than other state-of-the-art algorithms, and as demonstrated in the Unitree test case, maintain a similar control clock timing. With these results, we demonstrated that SOPPI effectively increases the algorithm's performance and/or particle efficiency over other methods, creating a viable high DOF control strategy.

In the future, we intend to expand this algorithm twofold.
First, we aim to explore further the multi-modal action search, identifying modalities explicitly, rather than implicitly, to further improve sample efficiency and representation. 
Secondly, we aim to integrate this work more directly into the Unitree's gait control and test it on real-world systems. Instead of relying on a single policy, we hope to implement multiple gaits (e.g., standing, walking, running, climbing stairs) and demonstrate abilities to switch between them, which would also translate well to real-world applications.



\bibliographystyle{IEEEtran}
\bibliography{references}

\end{document}